\documentclass[runningheads]{llncs}
\usepackage{graphicx}
\usepackage{comment}
\usepackage{amsmath,amssymb} 
\usepackage{color}
\usepackage{epsfig}
\usepackage{graphicx}
\usepackage{amsmath}
\usepackage{amssymb}
\usepackage{bm}
\usepackage{eucal}
\usepackage{multirow}
\usepackage{booktabs}
\usepackage{hyperref}

\makeatletter
\newcommand{\printfnsymbol}[1]{%
	\textsuperscript{\@fnsymbol{#1}}%
}
\makeatother

\begin{document}
\pagestyle{headings}
\mainmatter
\def\ECCVSubNumber{4148}  

\title{Fast and Accurate: Structure Coherence Component  for Face Alignment} 

\titlerunning{Fast and Accurate: Structure Coherence Component  for Face Alignment}
%
\author{Beier Zhu\inst{1}\thanks{equal contribution} \and
Chunze Lin\inst{1}\printfnsymbol{1} \and
Quan Wang \inst{1} \and
Renjie Liao\inst{2} \and 
Chen Qian \inst{1}}
\authorrunning{B. Zhu et al.}
%
\institute{SenseTime Research 
	\\ \email{\{zhubeier, linchunze, wangquan, qianchen\}@sensetime.com} \and
University of Toronto
\\ \email{rjliao@cs.toronto.edu}}
\maketitle
\begin{abstract}
	In this paper, we propose a fast and accurate coordinate regression method for face alignment.
	Unlike most existing facial landmark regression methods which usually employ fully connected layers to convert feature maps into landmark coordinate, we present a structure coherence component to explicitly take the relation among facial landmarks into account.
	Due to the geometric structure of human face, structure coherence between different facial parts provides important cues for effectively localizing facial landmarks. 
	However, the dense connection in the fully connected layers overuses such coherence, making the important cues unable to be distinguished from all connections. 
	Instead, our structure coherence component leverages a dynamic sparse graph structure to passing features among the most related landmarks.
	Furthermore, we propose a novel objective function, named Soft Wing loss, to improve the accuracy.
	Extensive experiments on three popular benchmarks, including WFLW, COFW and 300W, demonstrate the effectiveness of the proposed method, achieving state-of-the-art performance with fast speed. Our approach is especially robust to challenging cases resulting in impressively low failure rate ($0\%$ and $2.88\%$) in COFW and WFLW datasets.
%

%
%
\end{abstract}

\section{Introduction}
Face alignment, also known as facial landmark detection is an important topic in computer vision and has attracted much attention over past few years~\cite{lab,wing,DeCaFA,DVLN}.
As a fundamental step for face image analysis, face alignment plays a key role in many face applications such as face recognition~\cite{facerecogn1}, expression analysis~\cite{faceAU} and face editing~\cite{face2face}. 
Although significant progress has been made, face alignment is still a challenging problem due to issues like occlusion, large pose and complicated expression.

\begin{figure}[h]
	\centering
	\includegraphics[width=1\linewidth]{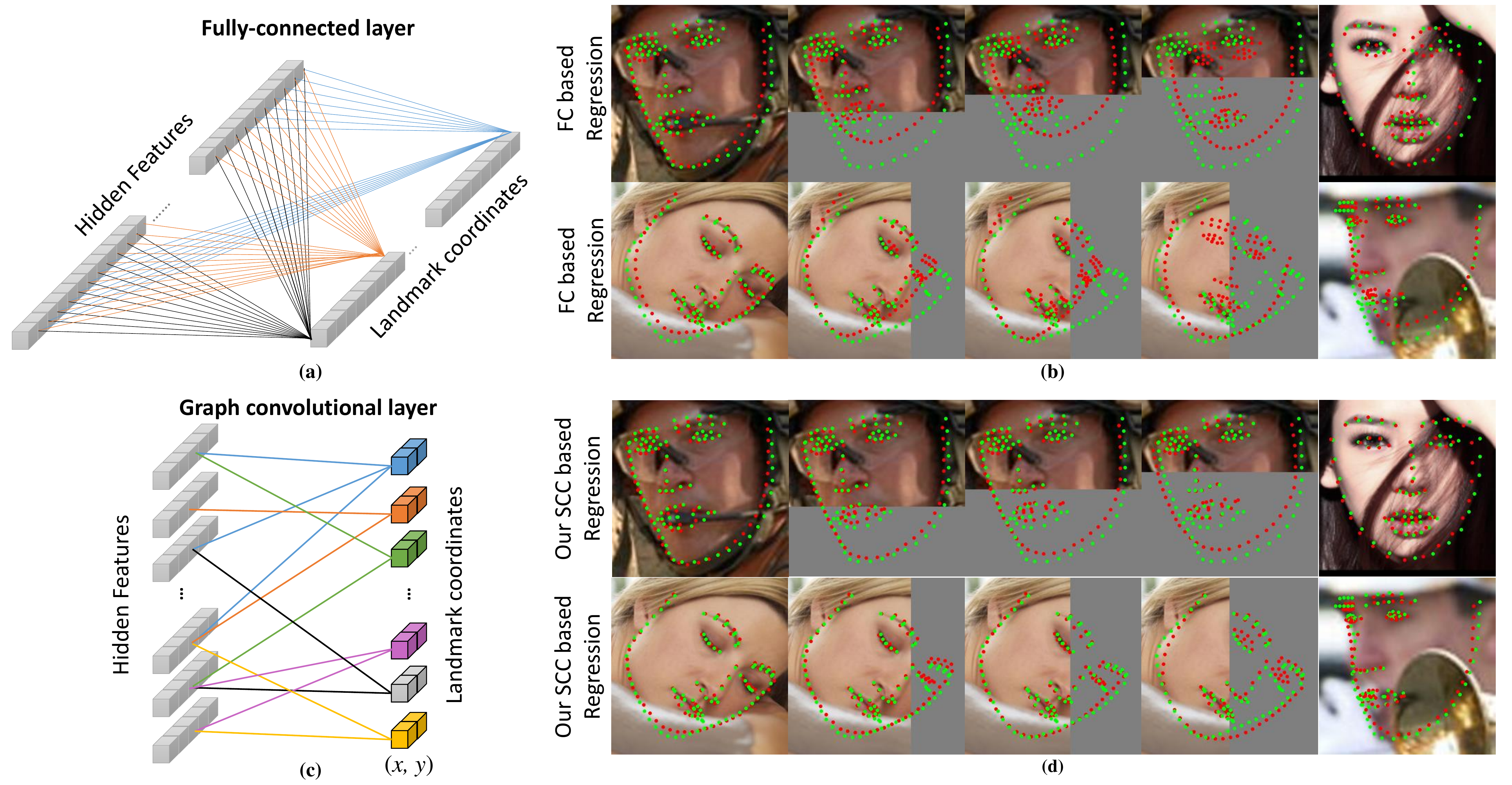}
	\caption{Comparison between the fully connected layer and our graph convolutional layer. (a) Dense connection in the fully connected layer. (b, d) The performance of fully connected based and our Structure Coherence Component based method under different levels of occlusions. Green and red points correspond to ground-truth and prediction, respectively.  (c) Sparse and relation-aware graph convolutional layer.}
	\label{fig:motivation}
\end{figure}

With the success of deep learning in several computer vision tasks such as image classification and object detection, many convolutional neural networks (CNN) based face alignment methods have been proposed.
Existing CNN-based face alignment methods can mainly be divided into two categories: coordinate regression based~\cite{Trigeorgis,wing,lab} and heatmap regression based ones~\cite{HGAlign,DeCaFA,HRNet}. 
Heatmap regression based methods commonly produce higher precise localization for its translation equivariant property~\cite{cohen2016group}.
Keeping the sizes of feature maps and heatmap is essential for high accuracy. However, it will also lead to computationally heavy models which are impractical for deployment in real-world applications.
{Coordinate regression based methods} are relatively simpler and can be built on lighter convolutional networks. 
The fully connected layers (FC) are commonly used in such methods to convert feature maps to facial landmark coordinates~\cite{Trigeorgis,wing,lab}.
However, the dense connections of fully connected layers make every landmark correlate to each other. 
As shown in Fig.~\ref{fig:motivation}(a), in the FC layer, every landmark coordinate is connected to the same hidden features.
The error of one landmark leads to error of all other landmarks, especially in hard cases such as occlusion. As shown in Fig.~\ref{fig:motivation}(b), when we progressively occlude human face, the error of face contour leads to the error of other parts of human face. 

\begin{figure*}[t]
	\centering
	\includegraphics[width=0.9\linewidth]{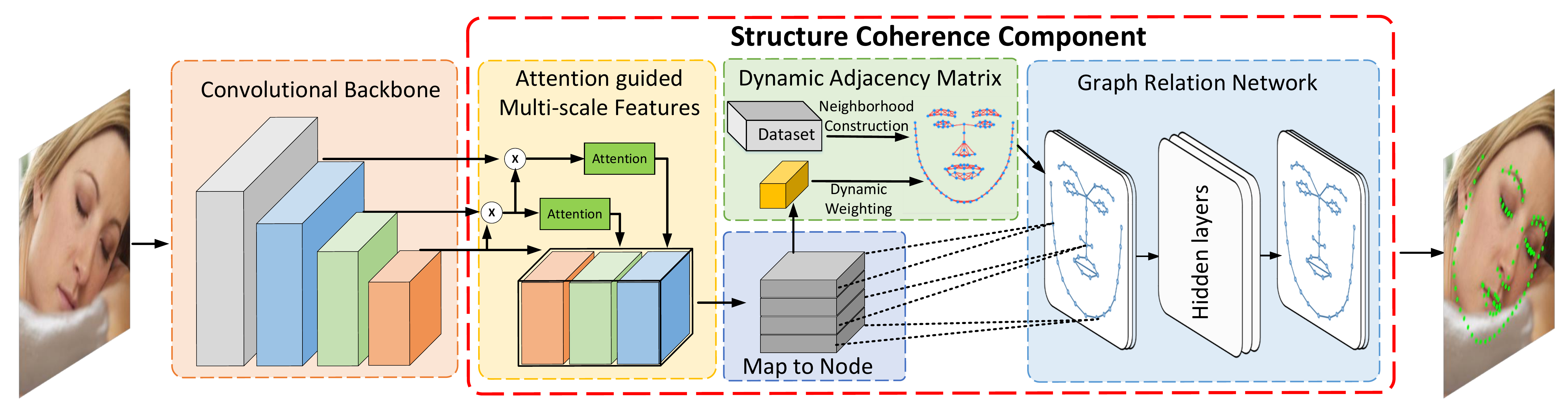}
	\caption{An overview of the proposed method. The convolutional backbone computes hierarchical feature maps from the input face image.
		These features are forward into our Structure Coherence Component which extracts spatial details and semantic information from different convolutional layers via attention. The map-to-node module then maps these attentive features into graph node representations. Together with the graph adjacency matrix that learned from the dataset and features extracted from map-to-node module, they are fed into the graph relation network to infer the facial landmarks.}
	\label{fig:pipeline}
\end{figure*}

Structure coherence between different facial parts provides important cues for effectively localizing facial landmarks, which helps keep the structure of face and predict occluded landmarks. 
In this paper, we propose \textbf{Structure Coherence Component (SCC)} to convert feature maps to facial landmark coordinates by explicitly exploring the relation among facial landmarks. With the help of deep geometric learning, we treat the intermediate features of each landmark as a node, and leverage a sparse graph structure to propagate features among the neighboring nodes, see Fig.~\ref{fig:motivation}(c). The sparse graph structure endows the model with the the capability of using the facial structure coherence appropriately. The sparse graph structure is learnt by data-driven based neighborhood construction and dynamic weight adjustment.
Fig.~\ref{fig:motivation}(d) shows that reasoning with structure coherence cues allows our model to correctly localize the key points in challenging real-world situations such as occlusion and large pose. 
As shown in Fig.~\ref{fig:pipeline}, Structure Coherence Component consists of four parts: attention guided multi-scale feature fusion, mapping to node,  dynamic adjacency matrix weighting module and graph relation network.  
The attention guided multi-scale feature fusion provides rich spatial details and semantic information features.
The mapping to node module converts these convolutional features into graph node representations and the relation is learnt via dynamic adjacency matrix weighting module, based on which, the graph relation network effectively regresses the coordinate of facial landmarks.
The proposed SCC, simple yet effective, permits more precise localization without burdening the model.

Furthermore, we propose Soft Wing Loss to handle the side-effect of Wing loss~\cite{wing} on small range errors.
Since the facial landmarks are not strictly defined, the annotations vary among annotators, introducing some shifts~\cite{SemAlign}.
In such a case, forcing the model to fit the ground-truth with a large gradient would cause unstable training.
Therefore, we make the model more focus on the errors of medium ranges.

We evaluate the proposed method on three widely-used face alignment benchmarks including  WFLW~\cite{lab}, COFW~\cite{RCPR} and 300W~\cite{300W}. Experimental results demonstrate the effectiveness of our approach, which outperforms existing state-of-the-art regression based methods by a large margin. 
In addition to the great performance, our model is much faster and lighter than the closest competitors.
We conduct extensive ablation studies to show the effectiveness of each proposed modules.

\section{Related Work}
\textbf{Traditional models:} Traditional facial landmark detection models mainly fall into two categories, i.e, fitting models and constrained local models. Taylor \textit{et al.} introduce the Active Appearance Model (AAM) \cite{aam1}\cite{aam2} to fits the facial images with a small number of coefficients, controlling both the facial appearance and the facial shape. Constrained local models\cite{cls1}\cite{cls2} are introduced to predict the landmarks based on the global facial shape constraints as well as the independent local appearance information around each landmark.
Locating facial landmarks with graph structure is related to some previous works\cite{ghiasi2014occlusion}\cite{zhu2012face}\cite{valstar2010facial}, which apply deformable part models (DPM)\cite{felzenszwalb2009object} to face analysis.
These methods belong to probabilistic graphical models, which require hand-crafted potential functions and iterative optimization for inference. 
However, our method is deep learning based graph network, which generates richer and more expressive feature embeddings and enjoys the faster inference.

\textbf{CNN based coordinate regression models:} Coordinate regression models directly map the face image to the landmark coordinates.
Zhang~\textit{et al.}~\cite{TCDCN} improve the robustness of detection through multi-task learning, i.e., learning landmark coordinates and predicting facial attributes at the same time. 
Feng~\textit{et al.}~\cite{wing} introduce a modified log loss, named Wing loss, to increase the contribution of small and medium errors to the training process.
LAB~\cite{lab} regresses facial landmark coordinates with the help of boundary information to reduce the annotation ambiguities.
In spite of the advantage of explicit inference of landmark coordinates without any post-processing, the coordinate regression models generally underperform heatmap regression models.

\textbf{CNN based heatmap regression models:}  Heatmap regression models leverage fully convolutional networks (FCNs) to maintain structure information throughout the whole network, and therefore outperform coordinate regression models.
In recent work, stacked hourglass (HG)~\cite{HG} is widely used to achieve the state-of-the-art performance.
Yang~\textit{et al.}~\cite{HGAlign} first normalize faces with a supervised transform and then prediction heatmap using a HG.
Liu~\textit{et al.}~\cite{SemAlign} develop a latent variable optimization strategy to reduce the impact of ambiguous annotations when training a 4-stacked HG. 
In addition to HG, architecture like HR-Net~\cite{HRNet} is also able to yield excellent performance.  
Despite their higher accuracy, heatmap regression models are much more costly from a computational point of view compared to coordinate regression models.

\textbf{Graph Neural Networks (GNNs):}
GNNs are a class of models which try to generalize deep learning to handle graph-structured data.
They are first introduced in~\cite{graphnn} and become more and more popular recently~\cite{geometricLecun}.
There are mainly two types of GNNs:
Message Passing based Neural Networks~\cite{graphnn,li2015gated,gilmer2017neural} and
Graph Convolution based Neural Networks~\cite{bruna2013spectral,SemiGraph,liao2019lanczosnet}. 
Many recent works have shown that GNNs are very effective in many computer vision tasks, e.g., RGBD semantic segmentation~\cite{qi20173d}, visual situation recognition~\cite{li2017situation}, scene graph generation and reasoning~\cite{yang2018graph,Shi_2019_CVPR},
image annotation~\cite{Zhang_2019_CVPR}, object detection~\cite{Xu_2019_CVPR} and
3D shape analysis\cite{feastnet}.
Specifically, in this work, we closely follow the so-called graph convolutional network (GCN)~\cite{SemiGraph} which greatly simplifies the graph convolution operator by exploiting approximation to the Chebyshev polynomial based graph spectral filters.
It provides a simple yet effective way to integrate local neighboring node feature following the graph topology. 


\section{Approach}
In this section, we present the proposed method in detail.
As illustrated in Fig.~\ref{fig:pipeline}, our Structure Coherence Component is mainly composed of four key parts: an attention guided multi-scale features fusion, a mapping to node module, dynamic adjacency matrix weighting module and a graph relation network.
Given an input face image, the convolutional backbone computes feature maps of different resolutions which are carefully fused via attention guidance.
A sparse graph structure is learnt by dynamic adjacency matrix weighting module.
The features extracted from attention module are then mapped into graph node representation and fed into the graph relation network which outputs the coordinates of facial landmarks.

\subsection{Attention Guided Multi-scale Features}
\begin{figure}[t]
	\centering
	\includegraphics[width=0.5\linewidth]{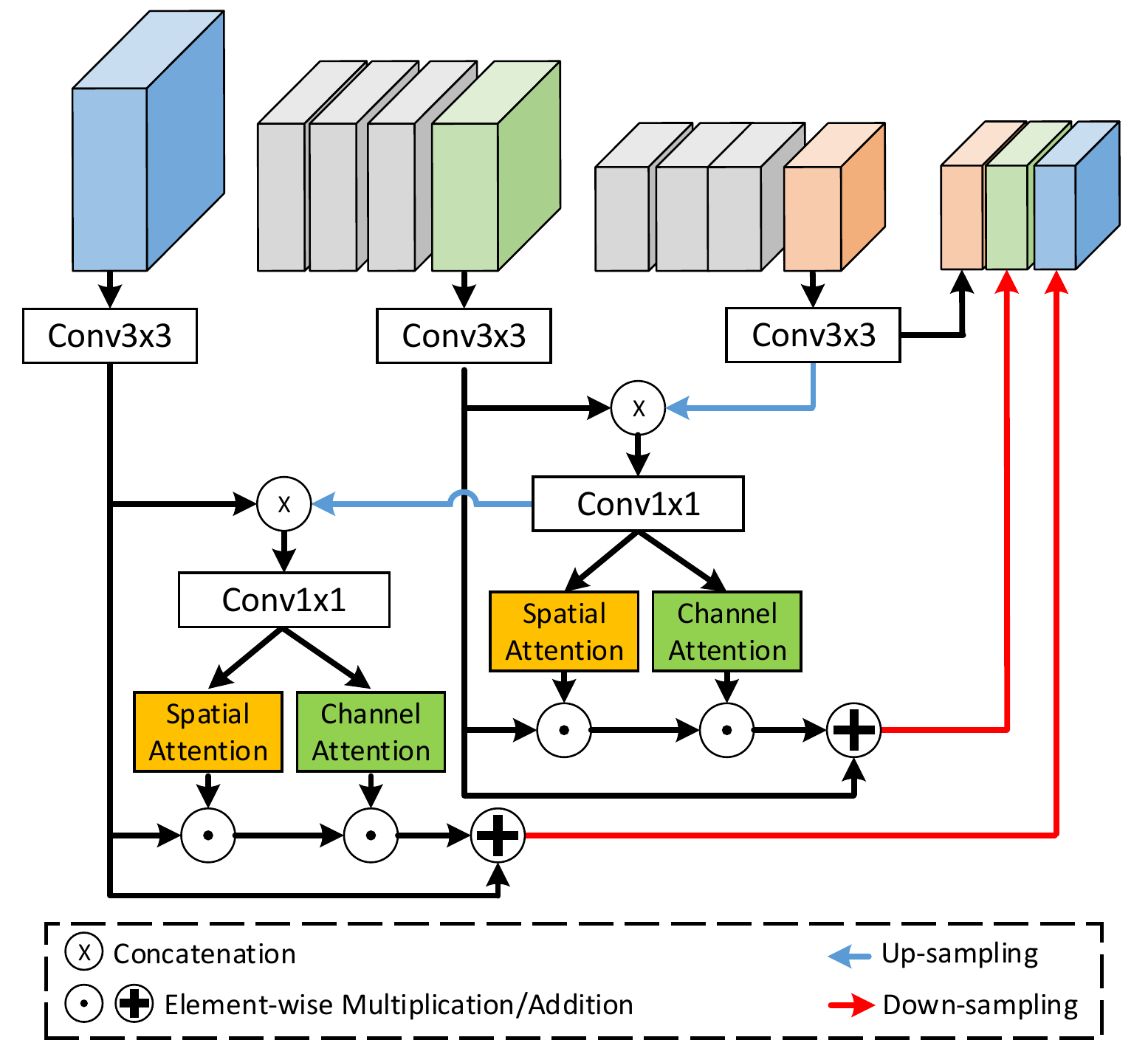}
	\caption{The architecture of the attention guided multi-scale feature learning module. We propagate the semantic information from deeper layers to guide shallower layers generating effective attention maps. After filtering out the noisy information with attention, we propagate spatial details from these shallower layers into the final feature maps.}
	\label{fig:multifeat}
\end{figure}

Since facial landmarks detection requires extreme precise localization, preserving the spatial information are crucial for an accurate model.
Heatmap based methods usually uses several hourglass structures~\cite{HG} to preserve the spatial information. 
However, such encoder-decoder architecture is extremely heavy and slows down the inference speed.
We propose an efficient attention guided multi-scale features module to improve the localization capability.
Fig.~\ref{fig:multifeat} illustrates the architecture of this module.

\textbf{Multi-scale Features:}
The feature maps from shallower layers encode low-level information and spatial details, while deep layers encode high semantic information~\cite{MSCNN,FPN,Linchunze}.
We introduce two bottom-up branches to propagate the spatial details from shallow layers to the deepest layer.
Specifically, consider a convolutional backbone composed of $L$ convolutional blocks. 
We denote $\bm{F}_l$ as the last feature maps of the $l$-th block.
We exploit the spatial information from the feature maps $\bm{F}_{L-1}$ and $\bm{F}_{L-2}$ to augment the localization precision of the features $\bm{F}_L$.
Each branch is composed of a $3 \times 3$ Conv-BN-ReLU, an attention mechanism to filter out noisy information and a down-sampling operation. 
These feature maps with spatial details are then concatenated with $\bm{F}_L$ to form more expressive feature maps.

\textbf{Semantic-guided Attention:}
Although the feature maps from shallow layers have rich spatial information, they also contain noisy information which are not informative from the perspective of semantic meaning.
We propose semantic-guided attention module to filter out such information.
Unlike existing self-attention which uses self-features to compute an attention map, we exploit the high-semantics features maps $\bm{F}_L$ to guide the feature maps $\bm{F}_{L-1}$ to suppress noisy information while keeping spatial details.
We first upsample the feature maps $\bm{F}_L$, concatenate it with $\bm{F}_{L-1}$ and reduce the channel dimension into the channel of $\bm{F}_{L-1}$ via an $1\times 1$ convolution, obtaining $\tilde{\bm{F}}_{L-1}$.
As $\tilde{\bm{F}}_{L-1}$ merges the information from $\bm{F}_L$ and $\bm{F}_{L-1}$, it contains both semantic information and spatial details.
We then use the attention module described in~\cite{cbam} to generate spatial attention $\bm{A}^s$ and channel-wise attention $\bm{A}^c$ from $\tilde{\bm{F}}_L$ as:
\begin{align} 
\label{eq:attention}
\bm{A}^c & = \sigma(\mathbf{W}_{1}\rho(\mathbf{W}_{0} \tilde{\bm{F}}^c_{\text{avg}}) + \mathbf{W}_{1}\rho(\mathbf{W}_{0} \tilde{\bm{F}}^c_{\text{max}})) \nonumber \\
\bm{A}^s & = \sigma(\text{conv}_{7\times7}([\tilde{\bm{F}}^s_{\text{avg}};\tilde{\bm{F}}^s_{\text{max}}])))\text{,}
\end{align}
where $\mathbf{W}_{0} \in \mathbb{R}^{C/2 \times C}$ and $\mathbf{W}_{1}\in \mathbb{R}^{C \times C/2}$,  $C$ is the channel number, $\text{conv}_{7\times 7}$ denotes an $7\times 7$ convolution operation, $\sigma$, $\rho$ and $[\cdot; \cdot]$ denote sigmoid function, ReLU activation and concatenation operation respectively, $\tilde{\bm{F}}^c_{\text{avg}}$/$\tilde{\bm{F}}^s_{\text{avg}}$ and $\tilde{\bm{F}}^c_{\text{max}}$/$\tilde{\bm{F}}^s_{\text{max}}$ denote spatially/channel-wise average-pooled features and max-pooled features, respectively. 
The notation $L$ is omitted for more clarity.
The attentive features are then obtained via element-wise multiplication and residual addition.
Similarly, we compute the attention for feature maps $\tilde{\bm{F}}_{L-2}$ with messages from features $\tilde{\bm{F}}_{L-1}$ and $\tilde{\bm{F}}_{L}$, and obtain the attentive features.
Finally, we concatenate these features to form the attention guided multi-scale features $\bm{F}_A$.
Note that designing the attention module is not our main focus, we adopt the commonly used attention module~\cite{cbam} in our semantic-guided process.

\subsection{Graph Relation Network}
As the relative spatial relationship of facial landmarks is stable, it is desirable to capture and exploit such important cues.
We statically calculate the correlation between face landmarks from data analysis and leverage graph relation networks to effectively explore these relation information. 

\textbf{Map-to-Node Module:} 
In order to make our network end-to-end trainable, we design the map-to-node
module to seamlessly map convolutional feature maps to graph node representations.
The input convolutional feature maps $\bm{F}_A\in \mathbb{R}^{C\times H \times W}$ (where $C$, $H$ and $W$ represents number of channels, height and width) are first transformed to the hidden feature maps by non-linear function $\bm{Z}=\phi(\bm{F}_A)\in \mathbb{R}^{Nn\times H\times W}$, where $n\in \mathbb{Z}^{+}$ is the expansion coefficient and $N$ is the number of landmarks.  
In this paper, we consider two convolution-BN-ReLU blocks with $n=4$ as the non-linear function $\phi(\cdot)$.
$\bm{Z}$ is then reshaped to $\bm{Z}_{\mathrm{node}} \in \mathbb{R}^{N\times nHW}$ to represent the node feature.

\textbf{Graph Convolution:} Unlike standard convolutions that operate on local Euclidean structures, e.g., a image grid, the goal of GCN is to learn a function $f(\cdot,\cdot)$ on a graph $\CMcal{G}$ , which takes node feature $\bm{H}^{l}\in \mathbb{R}^{N \times d_l}$ and the corresponding adjacency matrix $\bm{A}\in \mathbb{R}^{N \times N}$ as input, and outputs the node features as $\bm{H}^{l+1}\in \mathbb{R}^{N \times d_{l+1}}$. 
Here $N$, $l$, $d_l$ and $d_{l+1}$ denote the number of nodes, layer index, the dimension of input node features and the dimension of output node feature respectively.
Every GCN layer can be written as a non-linear function by,
\begin{equation}\label{Eq:BaseGCN}
\bm{H}^{l+1}=f(\bm{H}^l,\bm{A})
\end{equation}
With the specific graph convolutional operators employed by \cite{SemiGraph}, the layer can be represented as,
\begin{equation}
\bm{H}^{l+1}=\psi (\widetilde{\bm{D}}^{-\frac{1}{2}}\widetilde{\bm{A}}\widetilde{\bm{D}}^{-\frac{1}{2}}\bm{H}^{l}\bm{W}^{l})
\label{equ:gcnconv}
\end{equation}
where $\bm{W}^l\in \mathbb{R}^{d_l\times d_{l+1}}$ is a transformation matrix to be learned, $\widetilde{\bm{A}}=\bm{A}+\bm{I}$, $\widetilde{\bm{D}}$ is the degree matrix of $\widetilde{\bm{A}}$, $\widetilde{\bm{D}}^{-\frac{1}{2}}\widetilde{\bm{A}}\widetilde{\bm{D}}^{-\frac{1}{2}}$ is the symmetric normalized version of $\widetilde{\bm{A}}$ and $\psi(\cdot)$ denotes BN-ReLU operation.

\textbf{Neighborhoods Construction:} 
Graph relation network propagates information between nodes based on the adjacency matrix which is crucial to be correctly constructed. 
In our problem, due to the lack of pre-defined adjacency matrix for facial landmarks, we build it through a data-driven way, \textit{i.e.}, treating each landmark as a node and mining the correlation between landmarks within the dataset. 
Specifically, we assemble the landmark coordinates of the dataset into a rank-three data tensor $\bm{T}\in \mathbb{R}^{M\times N\times 2}$ where $M$ is the number of images, and the last dimension represents the $(x,y)$ coordinates.
We then slice the tensor $\bm{T}$ along the last dimension to generate $\bm{T}_x$ and $\bm{T}_y$. Based on $\bm{T}_x\in \mathbb{R}^{M\times N}$ and $\bm{T}_y\in \mathbb{R}^{M\times N}$, we calculate Pearson's correlation coefficient in $x$ and $y$ direction respectively to form correlation matrices $\bm{C}_x \in \mathbb{R}^{N \times N}$ and $\bm{C}_y \in \mathbb{R}^{N \times N}$.
Then, the correlation between nodes is defined as:
\begin{equation}
\bm{C} = \frac{1}{2}(\text{abs}(\bm{C}_x) + \text{abs}(\bm{C}_y))
\end{equation}
where $\text{abs}(\cdot)$ returns element-wise absolute value of matrix. 
Considering the computation cost and noisy edges, we only retain the top $k+1$ largest value of each row of $\bm{C}$ to form a sparse adjacency matrix $\bm{M}$. 
In other words, most $k$ relevant landmarks are picked as the neighborhood of each landmark. The binary adjacency matrix with self-loops can be written as:
\begin{equation}
\bm{M}_{ij}=\left\{
\begin{array}{ll}
1,\ &\text{if}\ \bm{C}_{ij} \in \text{Top}^{k+1}_{t=1,...,N}(\bm{C}_{it}) \\
0,\  & \text{otherwise}
\end{array}
\right.
\label{equ:topk}
\end{equation}


\textbf{Dynamic Adjacency Matrix Weighting:} 
The static adjacency matrix $\bm{M}$  is constructed based on the geometric structure of facial landmarks, while learning
relationship among landmarks for each face aims to take the facial appearance factors like occlusion and head pose into consideration. Given the binary matrix $\bm{M}$ which determines the node neighborhoods, we seek to adaptively adjust its weights.

Formally, given the features $\bm{Z}$ extracted from the map-to-node modules, we use the global average pooling layer followed by two fully connected layers to 
map $\bm{Z}$ to vector $\bm{a}$ whose size is equal to the non zeros in $\bm{M}$. Finally, we replace the non zeros value in $\bm{M}$ with $\bm{a}$ to form the dynamic adjacency matrix $\bm{A}$. 
Following the strategy in \cite{zhaoCVPR19semantic}, we adopt a row-wise softmax operation $\sigma_\text{row}$ to replace the symmetric normalization in Eq.~\ref{equ:gcnconv}.
Softmax operation makes the weights of each node like probabilities over its neighboring nodes, which stabilizes the training process:
\begin{equation}\label{Eq:learnableMatrix}
\bm{H}^{l+1}=\psi(\sigma_\text{row}(\bm{A})\bm{H}^l\bm{W}^l)
\end{equation}

We use the binary matrix $\bm{M}$ to hold the neighborhoods and only learn their weights because the facial shape pattern are stable, 
fix the sparse connection will greatly reduce the training parameters which makes the learning process easier.




\textbf{Graph Relation Network:}
Inspired by the success of ResNet\cite{ResNet}, we adopt the graph residual block architecture. Each block consists of two graph convolutional layers and can be formulated based on Eq.~(\ref{Eq:BaseGCN}) as
\begin{align}
\bm{H}^{l+1}&=f(\bm{H}^l,\bm{A}) \nonumber \\
\bm{H}^{l+2}&= f(\bm{H}^{l+1},\bm{A}) + \bm{H}^l
\end{align}
The overall graph relation network architecture is shown in Fig.~\ref{fig:pipeline}.
The input feature $\bm{Z}_{node}$ is first fed to graph convolution, followed by several graph residual blocks. 
The last graph convolution (without batch normalization and ReLU) block maps the hidden node features to landmark coordinates $\bm{O}\in\mathbb{R}^{N\times 2}$.

\textbf{Comparison with FC-based regression methods.} 
The fully connected layer and our graph convolutional layers embed the feature of landmarks in two different ways.
As shown in Fig.\ref{fig:motivation}(a) The CNN backbone and the hidden fully connected layer map the input facial image to the hidden vector,
 which embed the feature of landmarks globally.
Thus, the errors of some parts of the prediction effects the other parts, as they share the same hidden feature.
As shown in Fig. \ref{fig:motivation}(b), for the FC-based method, the errors of occluded part interfere the prediction of other visible parts.
Meanwhile, our SCC embeds the node feature for each landmark, and propagates node feature according to their relationship.
If some parts of predictions fail because of the occlusion, large pose or other hard condition, the node feature of other parts degrade gracefully because of the sparse connection among the node features and the dynamic adjustment of the relationship.   
As shown in Fig. \ref{fig:motivation}(d), the SCC-based method are more robust to hard cases. 
Besides, fully connected layers are prone to overfit because of the large number of trainable parameters, while the graph convolution layer requires fewer trainable parameters.

\subsection{Soft Wing Loss}
\label{sec:softwing}

\begin{figure}[t]
	\centering
	\includegraphics[width=0.45\linewidth]{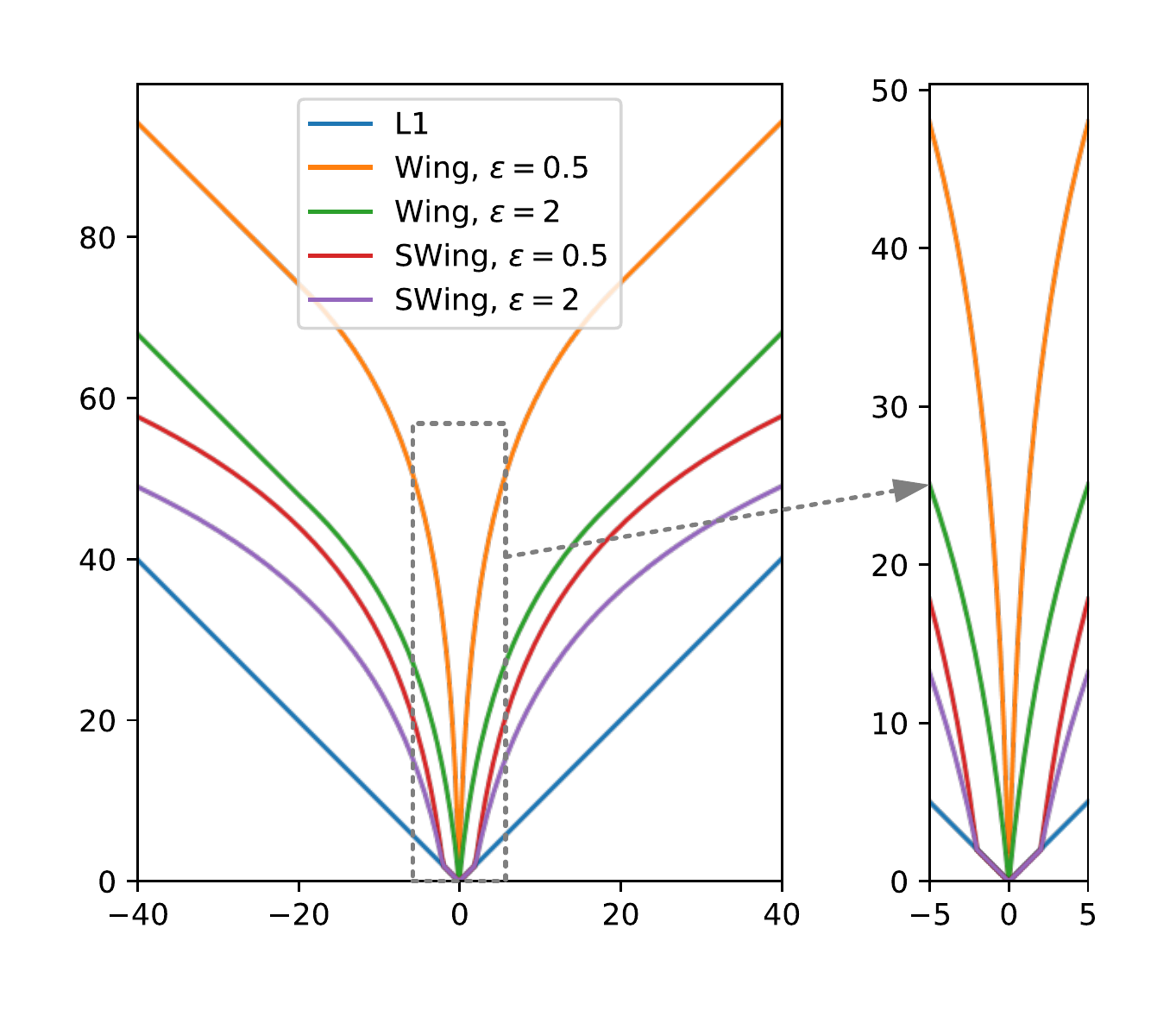}
	\caption{Illustration of L1, Wing and Soft Wing loss functions. $\omega_1$ is set 2. $\omega$ and $\omega_2$ are set to 20. Unlike Wing loss, our loss is linear for small errors.}
	\label{fig:loss_demo}
\end{figure}

Wing loss\cite{wing} has constant gradient when error is large, and large gradient for small or medium range errors, which is defined as:
\begin{equation}
\mathrm{Wing}(x)=\left\{
\begin{array}{ll}
\omega \ln({1+\frac{|x|}{\epsilon}})& \mathrm{if}\ |x| < \omega \\
|x| - C  &\mathrm{otherwise}
\end{array}
\right.
\end{equation}
where $x$ is error and $C$ is $\omega - \omega \ln (1+\omega/\epsilon)$ to smoothly link two piece-wise functions. 
According to our experiment, the performance of Wing loss is not consistently better than L1 loss, especially when we train the neural networks on difficult dataset with heavy occlusion and blur, such as WFLW.
As mentioned in \cite{SemAlign}, this may be caused by inconsistent annotations due to various reasons, e.g., unclear or inaccurate definition of some landmarks, poor quality of some facial images.
Imposing a large gradient magnitude around very small error to force the model exactly fit the ground truth landmarks makes the training process unstable. 
To alleviate this problem, we present Soft Wing loss to more focus on the errors of medium range:
\begin{equation}
\text{SoftWing}(x)=\left\{
\begin{array}{ll}
|x| &\text{if}\ |x| < \omega_1 \\
\omega_2 \ln({1+\frac{|x|}{\epsilon}})+B  &\text{otherwise}
\end{array}
\right.
\end{equation}
which is linear for small values, and take the curve of $\ln (\cdot)$ for medium and large values. Similar to Wing loss, we use the non-negative $\omega_1$ to switch between linear and non-linear part, and $\epsilon$ to limit the curvature of the non-linear part. $B$ is set to $\omega_1-\omega_2\ln (1+\omega_1/\epsilon)$ to make function continuous at $\omega_1$. The visualization of L1, Wing and our Soft Wing loss is shown in Fig.\ref{fig:loss_demo}. 
Note that we discard the linear part of Wing loss, since our proposed loss can adaptively adjust the magnitude of gradient between medium ($\omega_1<|x|<\omega_2$) and large errors ($|x|> \omega_2$). The magnitude of gradient of the non-linear part is $\frac{\omega_2}{|x|+\epsilon}\approx \frac{\omega_2}{|x|}$ ($\epsilon$ is commonly set to small value). Our proposed loss is insensitive to outliers where the gradient varies between $[\frac{\omega_2}{C},1]$ ($C$ is the image size). Note that $\omega_2$ should not set to small value because it will cause gradient vanishing problem.


\begin{table*}[t]
	\centering
	\scalebox{0.8}{
		\begin{tabular}{c|c|c|c|c|c|c|c|c}
			\hline
			Metric                      & Method            & Fullset & Pose  & Expression & Illumination & Make-up & Occlusion & Blur\\ \hline\hline
			\multirow{8}{*}{NME}   & DVLN\tiny{17}\small{\cite{DVLN}}     & 6.08 & 11.54  & 6.78 & 5.73 & 5.98 & 7.33 & 6.88 \\
			& LAB\tiny{18} \small{\cite{lab}}       & 5.27 & 10.24  & 5.51 & 5.23 & 5.15 & 6.79 & 6.32\\
			& Wing\tiny{18} \small{\cite{wing}}     & 5.11 & 8.75   & 5.36 & 4.93 & 5.41 & 6.37 & 5.81 \\ 
			& AGCFN\tiny{19} \small{\cite{AGCFN}}        & 4.90 & 8.78   & 5.00 & 4.93 & 4.85 & 6.26 & 5.73 \\ 
			& LAB\tiny{18} \small{\cite{lab}} + AVS\tiny{19} \small{\cite{AVS}} & 4.76 & 8.21 & 5.14 & 4.51 & 5.00 & 5.76 & 5.43 \\
			& DeCaFA\tiny{19} \small{\cite{DeCaFA}} & 4.62 & 8.11   & \textbf{4.65} & 4.41 & 4.63 & 5.74 & 5.38 \\ 
			& HRNet\tiny{19} \small{\cite{HRNet}}        & 4.60 & 7.94   & 4.85 & 4.55 & \textbf{4.29} & 5.44 & 5.42 \\ 
			& \textbf{Ours}                                       & \textbf{4.40} & \textbf{7.52}   & \textbf{4.65} & \textbf{4.31} & 4.36 & \textbf{5.23} & \textbf{5.04} \\    \hline
			\multirow{6}{*}{FR}    & DVLN\tiny{17}\small{\cite{DVLN}}     & 10.84& 46.93  & 11.15& 7.31 & 11.65& 16.30 & 13.71  \\  
			& LAB\tiny{18} \small{\cite{lab}}       & 7.56 & 28.83  & 6.37 & 6.73 & 7.77 & 13.72 & 10.74 \\
			& Wing\tiny{18} \small{\cite{wing}}     & 6.00 & 22.72  & 4.78 & 4.30 & 7.77 & 12.50 & 7.76 \\ 
			& AGCFN\tiny{19} \small{\cite{AGCFN}}        & 5.92 & 24.23  & 5.41 & 4.72 & 5.82 & 11.00 & 8.79 \\     
			& LAB\tiny{18} \small{\cite{lab}} + AVS\tiny{19} \small{\cite{AVS}} & 5.24 & 20.86  & 4.78 & 3.72 & 6.31 & 9.51  & 7.24 \\ 
			& DeCaFA\tiny{19} \small{\cite{DeCaFA}} & 4.84 & 21.4   & 3.73 & 3.22 & 6.15 & 9.26  & 6.61 \\
			& \textbf{Ours}                                       & \textbf{2.88} & \textbf{13.80}  & \textbf{2.55} & \textbf{2.29} & \textbf{2.43} & \textbf{5.98}  & \textbf{4.14}\\\hline
			\multirow{6}{*}{AUC}        & DVLN\tiny{17}\small{\cite{DVLN}}     & 0.4551  & 0.1474 & 0.3889 & 0.4743 & 0.4494 & 0.3794 & 0.3973 \\
			& LAB\tiny{18} \small{\cite{lab}}       & 0.5323  & 0.2345 & 0.4951 & 0.5433 & 0.5394 & 0.4490 & 0.4630  \\
			& Wing\tiny{18} \small{\cite{wing}}     & 0.5504  & 0.3100 & 0.4959 & 0.5408 & 0.5582 & 0.4885 & 0.4918  \\ 
			& AGCFN\tiny{19} \small{\cite{AGCFN}}        & 0.5452  & 0.2826 & 0.5267 & 0.5511 & 0.5547 & 0.4621 & 0.4823  \\ 
			&LAB\tiny{18} \small{\cite{lab}} + AVS\tiny{19} \small{\cite{AVS}}  & 0.5460  & 0.2764 & 0.5098 & 0.5660 & 0.5349 & 0.4700 & 0.4923 \\ 
			& DeCaFA\tiny{19} \small{\cite{DeCaFA}} & 0.563   & 0.292  & \textbf{0.546}  & \textbf{0.579}  & \textbf{0.575}  & 0.485  & 0.494 \\
			& \textbf{Ours}                                       & \textbf{0.5666}  & \textbf{0.2981} & {0.5430} & {0.5761} & {0.5710} & \textbf{0.4936} & \textbf{0.5095} \\\hline
			
		\end{tabular}
	}
	\caption{Evaluation of our method and state-of-the-art approaches on Fullset and six typical subsets of WFLW. The results in terms of normalized mean error, NME~($\%$), failure rate at $10\%$, FR~($\%$) and AUC are reported.}
	\label{tab:WFLW}
\end{table*}

\section{Experiments}
In this section, we evaluate our method on three popular face alignment benchmarks, compare with state-of-the-art approaches and conduct the ablation study. 

\subsection{Experimental Setup}

\textbf{Datasets:} We conduct evaluation on three widely-adopted challenging datasets: WFLW~\cite{lab}, COFW~\cite{RCPR} and 300W~\cite{300W}.
\textbf{WFLW} is among the most challenging face alignment benchmark which includes various hard cases such as heavy occlusion, blur and large pose.
\textbf{COFW} is collected to present faces with large variations in shape and occlusions in real-world conditions.
Various types of occlusions are introduced and result in a $23\%$ occlusion on facial parts on average.
We also use the re-annotated test set~\cite{HPM} with 68 landmarks annotation for cross-dataset validation.
\textbf{300W} contains face images with moderate variations in pose, expression and illumination. 

\textbf{Evaluation Metric:}
We evaluate the proposed method with normalized mean error and failure rate. we use the inter-ocular distance as the normalization factor.
Following the protocol in \cite{lab}, the failure rate for a maximum error of 0.1 is reported.
Area under curve (AUC) is also calculated based on the cumulative error distribution for WFLW dataset.

\subsection{Implementation details}
All training images are center-cropped and resized to $256\times 256$.
Data augmentation is performed with random rotation ($\pm40^{\circ}$), translation ($\pm 30\ px$), flipping ($50\ \%$), rescaling ($\pm10\ \%$) and occlusion ($20\ \%$ of image size). 
To mitigate the issue of pose variations, we adopt the Pose-based Data Balancing (PDB)\cite{wing} strategy with 9 bins.
We use ResNet34\cite{ResNet} as our backbone.
During the training, we employ vanilla SGD for optimization with a batch size of $64$ for $500$ epochs.
We set the weight decay and the momentum to $0.0005$ and $0.9$ respectively.
The initial learning rate is $0.01$ which is dropped by 5 every $100$ epochs.
The parameters of the Soft Wing loss are set to $\omega_1=2$, $\omega_2=20$ and $\epsilon=0.5$.
The $k$ is set to 3 for adjacency matrix.
We use 4 graph residual blocks with hidden feature dimension 128.
Our models are trained from scratch using Pytorch.

\subsection{Comparison with the State of the Art}

\begin{figure}[t]
	\centering
	\includegraphics[width=0.8\linewidth]{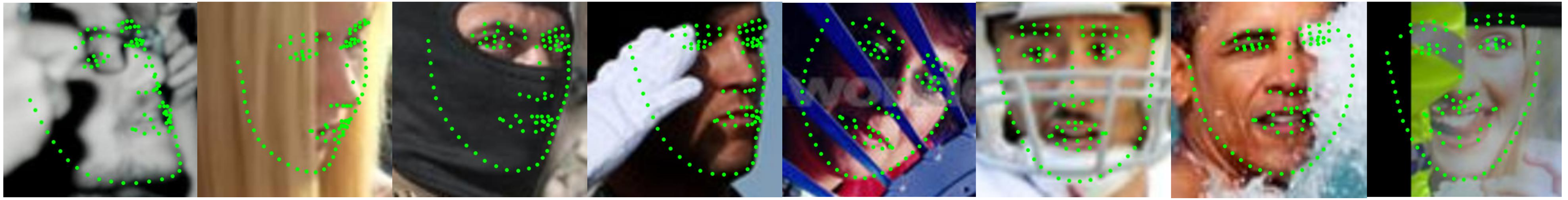}
	\caption{Visualization of some hard cases from the WFLW testset.}
	\label{fig:hard_case}
\end{figure}

\textbf{WFLW:}
We evaluate our approach on the WFLW dataset and compare with state-of-the-art methods in terms of mean error, failure rate and AUC.
To better understand the effectiveness of the proposed method, we analyse the performance on six subsets with specific issue, \textit{e.g.}, large pose, occlusion and exaggerated expression~\cite{lab}.
The overall results are tabulated in Table~\ref{tab:WFLW}.
The proposed method achieves $4.40\%$ NME, $2.88\%$ failure rate and $0.5666$ AUC, which outperforms most state-of-the-art approaches.
Our method fails on only $2.88\%$ of all images, which demonstrates the robustness of our model.
Qualitative results are depicted in Fig.~\ref{fig:hard_case}, where our model successively localizes landmarks in hard cases.

\begin{table}[t]
	\centering
	\scalebox{0.8}{
		\begin{tabular}{c|cc|cc}
			\hline
			\multirow{2}{*}{Method} & \multicolumn{2}{c|}{Trained on COFW} & \multicolumn{2}{c}{Trained on 300W} \\ \cline{2-5} 
			& NME       & FR    & NME     & FR                   \\ \hline
			TCDCN\tiny{14}\small{\cite{TCDCN}}  & -     & -     & 7.66 & 16.17 \\
			SAPM\tiny{15}\small{\cite{BMVC2015_22}} &-  & -     & 6.64 & 5.72 \\
			CFSS\tiny{15}\small{\cite{CFSS}}    & -     & -     & 6.28 & 9.07 \\
			HPM\tiny{14}\small{\cite{HPM}}      & 7.50  & 13.00  & 6.72  & 6.71  \\
			CCR\tiny{15}\small{\cite{CCR}}      & 7.03  & 10.9   & -    & - \\
			DRDA\tiny{16}\small{\cite{DRDA}}    & 6.46  & 6.00   & -    & - \\
			RAR\tiny{16}\small{\cite{RAR}}      & 6.03  & 4.14   & -    & - \\
			SFPD\tiny{17}\small{\cite{SFPD}}    & 6.40  & -     & -     & - \\
			DAC-CSR\tiny{17}\small{\cite{DAC-CSR}} & 6.03  & 4.73 & -  & - \\
			Wing\tiny{18}\small{\cite{wing}}    & 5.44  & 3.37 & -    & -\\
			ODN\tiny{19}\small{\cite{ODN}}      & 5.30  & - & -  & - \\
			LAB\tiny{18}\small{\cite{lab}}      & 3.92  & 0.39  &4.62 & 2.17 \\ 
			SAN\tiny{18}\small{\cite{SAN}} + AVS\tiny{19}\small{\cite{AVS}} &-&-& 4.43 & 2.82 \\
			\hline \textbf{Ours}        & \textbf{3.63} & \textbf{0} &\textbf{4.18} & \textbf{0} \\
			\hline
	\end{tabular}}
	\caption{Evaluation on the COFW dataset in terms of NME ($\%$) and Failure Rate ($\%$) at $10\%$.}
	\label{tab:COFW}
\end{table}

\textbf{COFW:}
As shown in in Table~\ref{tab:COFW}, our method achieves state-of-the-art performance with $3.63\%$ mean error and $0\%$ failure rate.
To further verify the generalization capability of our method, we conduct a cross-dataset evaluation using COFW-68 dataset annotated with 68 landmarks~\cite{HPM}.
Our method outperforms the existing best approaches by a large margin, with $4.18\%$ mean error and $0\%$ failure rate.
Since the COFW dataset is mainly composed of occluded faces, this impressive performance indicates the robustness of our graph relation framework to handle heavy occlusions.


\textbf{300W:}
We compare our approach against the existing best performing methods on the 300W dataset.
The results are reported in Table~\ref{tab:300W}.
Our method outperforms most existing approaches.
Note that our method achieves the best results on the challenging subset, which highlights the robustness of the proposed approach in hard cases.

\begin{minipage}[t]{\textwidth}
 \begin{minipage}[t]{0.45\textwidth}
  \centering
     \makeatletter\def\@captype{table}\makeatother
     \scalebox{0.8}{
       \begin{tabular}{l|ccc}
			\hline
			Method    & Common    & Challenging    & Full   \\ \hline
			PCD-CNN\tiny{18}~\small{\cite{kumar2018disentangling}}   & 3.67      & 7.62    & 4.44 \\
			CPM+SBR\tiny{18}~\small{\cite{dong2018supervision}}   & 3.28      & 7.58    & 4.10 \\
			SAN\tiny{18}~\small{\cite{SAN}}       & 3.34      & 6.60    & 3.98 \\
			LAB\tiny{18}~\small{\cite{lab}}      & 2.98      & 5.19    & 3.49 \\
			DeCaFA\tiny{19}~\small{\cite{DeCaFA}}    & 2.93      & 5.26    & 3.39 \\
			HRNet\tiny{19} \small{\cite{HRNet}} &  \textbf{2.87} & 5.15 & 3.32 \\
			\hline 
			\textbf{Ours}      & 2.88    &\textbf{4.93}   & \textbf{3.28} \\
			\hline
	\end{tabular}}
	\vspace{-7pt}
	\caption{Evaluation on the 300W Common subset, Challenging subset and Fullset in terms of mean error($\%$).}
	\label{tab:300W}
  \end{minipage}
  \hspace{2pt}
  \begin{minipage}[t]{0.45\textwidth}
   \centering
        \makeatletter\def\@captype{table}\makeatother
        \scalebox{0.8}{
         \begin{tabular}{l|c c c}        
          \hline
			Model            & \# params (M) & FLOPS (G)   &  RT (ms)  \\ \hline
			SAN~\cite{SAN}   & 199.63           &    -        &   343      \\   
			LAB~\cite{lab}   & 32.05               &   28.583    &    60       \\
			Wing~\cite{wing} & 24.75            &  5.396      &   30  \\
			\textbf{Ours}    & 24.68            & \textbf{5.165}     &   \textbf{23}      \\ \hline
      \end{tabular}}
      \vspace{-7pt}
	\caption{Efficiency comparison in terms of number of parameters, FLOPS and runtime.}
	\label{tab:efficieny}
   \end{minipage}
\end{minipage}

\textbf{Efficiency Comparison:}
Since facial landmark detection is widely deployed for many real-time applications, the model size, FLOPS and processing speed are key criteria. 
We evaluate the runtime of our model on a 1080Ti GPU and compare with existing methods in Table~\ref{tab:efficieny}.
Our model only takes $23$ ms and $5.165$ FLOPS to process a $256 \times 256$ input image and consists of $\sim 24$M parameters.
Overall, our model is faster and smaller than most competitors.

\subsection{Ablation Study}
Our framework is composed of several pivotal modules such as graph relation network, attention guided multi-scale features, and soft-wing loss. 
Based on the baseline Resnet34 with $L=5$ layer stages, we examine the contributions of each proposed module on the WFLW dataset and report the overall results in Table~\ref{tab:ablation}.

\begin{minipage}{\textwidth}
 \begin{minipage}[t]{0.45\textwidth}
  \centering
     \makeatletter\def\@captype{table}\makeatother
       \scalebox{0.7}{
		\begin{tabular}{l|c c c c c c}
			\hline
			Component & \multicolumn{6}{c}{Choice} \\
			\hline
			Fully connected & \checkmark & & & &   &   \\
			GN &  & \checkmark & \checkmark & \checkmark &&  \checkmark\\
			Attention m-s F.  & & & \checkmark & & \checkmark & \checkmark\\
			Soft-Wing loss & & & & \checkmark & \checkmark  &\checkmark                 \\
			GN w/o DW&  &  & & & \checkmark & \\
			\hline
			NME ($\%$) & 5.95 & 4.64 & 4.53 & 4.52 & 4.47 & \textbf{4.40}\\
			\hline
	\end{tabular}}
	\caption{Ablation study on components on the WFLW dataset. GN: Graph Network. DW: Dynamic adjacency matrix Weighting}
	\label{tab:ablation}
    
  \end{minipage}
  \hspace{5pt}
  \begin{minipage}[t]{0.45\textwidth}
   \centering
        \makeatletter\def\@captype{table}\makeatother
       \scalebox{0.7}{
		\begin{tabular}{l|c}
			\hline
			Design Choice  & NME ($\%$) \\
			\hline
			Self-attention &    4.61         \\
			Semantic-guided attention & \textbf{4.53}  \\ 
			\hline
			Feature maps $\{\bm{F}_5\}$ &    4.64 \\
			Feature maps $\{\bm{F}_4,\bm{F}_5\}$ &    4.57 \\
			Feature maps $\{\bm{F}_3,\bm{F}_4,\bm{F}_5\}$ &  \textbf{4.53} \\
			Feature maps $\{\bm{F}_2,\bm{F}_3,\bm{F}_4,\bm{F}_5\}$ &  4.56  \\
			\hline
	\end{tabular}}
	\caption{Ablation study on attention generation methods and feature maps for spatial message propagation.}
	\label{tab:attention}
   \end{minipage}
\end{minipage}

%
%

\textbf{Baseline Model:} We first utilize the FC layers to directly regress the facial landmarks. This model is our baseline which achieves a NME of $5.95\%$.

\textbf{Graph Relation Network:}
The graph relation network is a key part of our Structure Coherency Component. 
We obtain a $1.31\%$ improvement by replacing the FC layers with our graph relation network, resulting in a NME of $4.64\%$.

\textit{Top-k value for adjacency matrix:}
We report the result with different values of $k$ from $k=1$ to $k=97$ in Fig. \ref{fig:topk_abla}. When $k=3$, our model achieves the best performance on WFLW dataset. Note that, the performance degrades if the adjacency matrix is too sparse or too dense. When $k$ is too small, each graph node can not get sufficient information from its correlated neighborhoods, meanwhile, when $k$ is too large, the adjacency matrix becomes dense which leads to oversmoothing of the node features.

\textit{Dynamic adjacency matrix weighting:} 
We replace the dynamic adjacency matrix with the binary adjacency matrix $\bm{M}$ and we observe a $0.07\%$ degradation.


\begin{table}[t]
\centering
	\scalebox{0.8}{
\begin{tabular}{l|lllllllll}
\hline
Top k & 1    & 2    & 3    & 4    & 5    & 10   & 20   & 40   & 97   \\ \hline
NME   & 4.44 & 4.43 & 4.40 & 4.46 & 4.50 & 4.59 & 4.69 & 4.71 & 4.75 \\ \hline
\end{tabular}}
\caption{NME(\%) comparison with different values of $k$. $k$ is for building adjacency matrix.}
\label{fig:topk_abla}
\end{table}


\begin{figure}[t]
	\centering
	\includegraphics[width=0.5\linewidth]{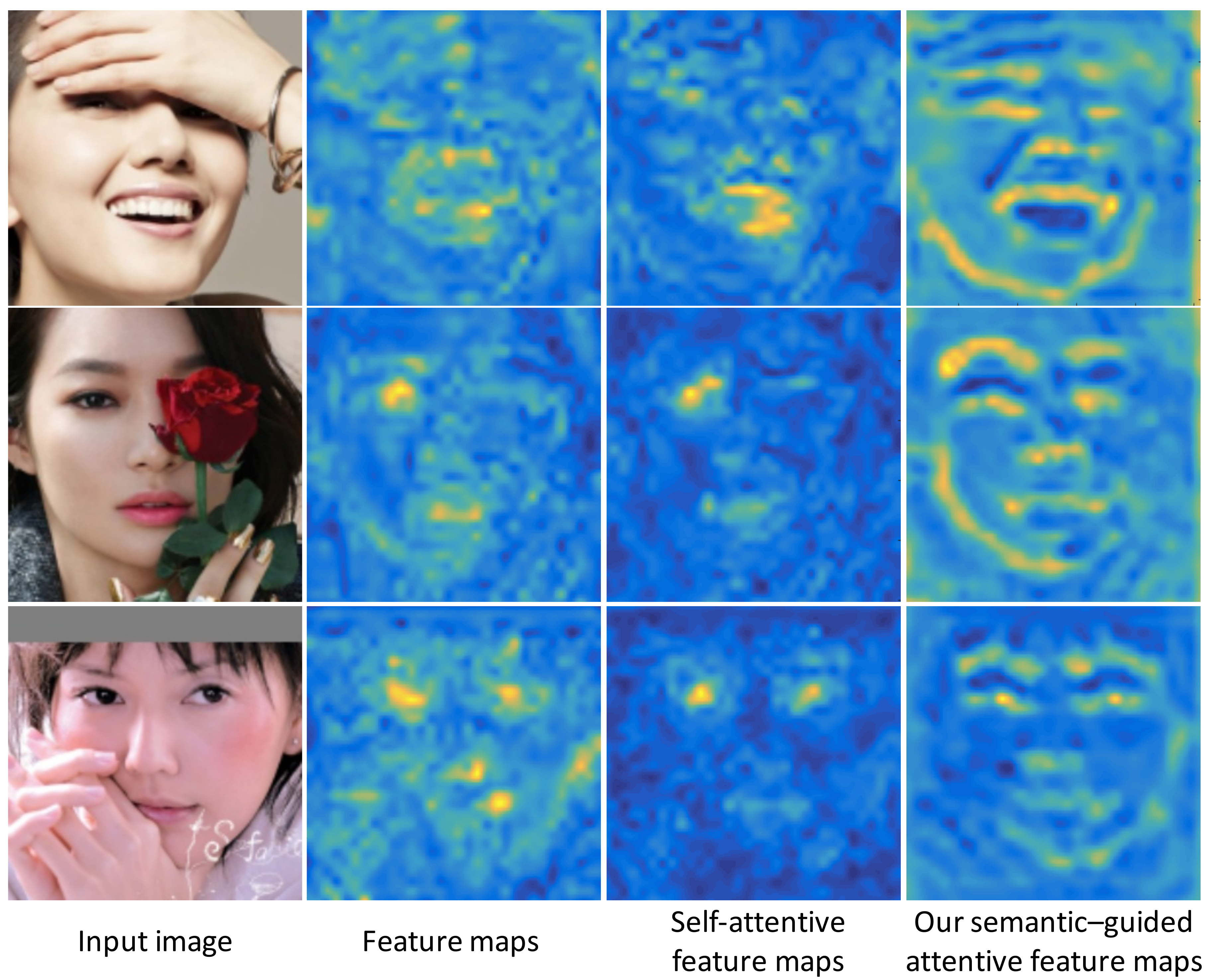}
	\caption{Qualitative analysis of the attention effects. The max along the channel of $\bm{F}_3$ is illustrated. Our semantic-guided attention highlights all visible key facial parts whereas the self-attention only highlights a few facial parts such as eyes or mouth.}
	\label{fig:attention_vis}
\end{figure}

\textbf{Attention guided multi-scale feature:} 
The attention guided multi-scale features fusion plays a key role in improving the representation capability of features. 
By endowing the spatial details to high-semantics features, the model performs a NME of $4.53\%$, which corresponds to a $0.11\%$ improvement. 

\textit{Semantic-guided attention}:
We examine the importance of incorporating additional semantic information from deeper layers to guide generating attention maps.
To this end, we degenerate our semantic-guided attention structure into a general self-attention mechanism.
As shown in Table~\ref{tab:attention}, we observe a drop of performance, resulting in a NME of $4.61\%$.
This experiment proves that the semantic information from high-level features is crucial to guide generating high quality attention.
The quantitative performances are supported by the qualitative results illustrated in Fig.~\ref{fig:attention_vis}.
The semantic guidance permits to make the feature maps focus on all visible key facial parts.
Since self-attention only explore self-information, it only highlights the high-activated part in feature maps.

\textit{Features combination:} 
To improve the localization capability, we propagate the spatial information from the shallower layers.
We study which combination of feature layers is the optimal one.
As tabulated in Table~\ref{tab:attention}, the performance increases with the the additional spatial information propagation and the combination of features $\{\bm{F}_3, \bm{F}_4, \bm{F}_5\}$ yields best results. 
Since layer 2 is quite shallow, $\bm{F}_2$ consists few useful information and limits the performance due to the noise.

\textbf{Soft Wing Loss:} 
The Soft Wing loss improves the results of the graph relation network by $0.12\%$.
We compare the performance of L1, Wing and our Soft Wing loss based on our baseline model, the results are shown in Table~\ref{tab:loss}. Our Soft Wing loss consistently outperforms Wing loss and L1 loss. As discussed in Section \ref{sec:softwing}, the performance of Wing loss degrades when $\epsilon$ decreases, while our loss benefits from imposing larger gradients on medium range errors. The performance of Wing loss is even worse than L1 loss when $\epsilon$ is very small.

%

\begin{table}[t]
\centering
	\scalebox{0.8}{
	\begin{tabular}{l|cccccc}
	\hline
epsilon & 0.1  & 0.2  & 0.5  & 1    & 1.5  & 2    \\ \hline
L1                      & \multicolumn{6}{c}{5.95}                \\
Wing                    & 6.52 & 5.98 & 5.81 & 5.78 & 5.75 & 5.72 \\
SoftWing                & 5.70 & \textbf{5.66} & 5.67 & 5.71 & 5.70 & 5.71 \\ \hline
\end{tabular}}
	\caption{Comparison of different loss functions.
		Analysis shows the effectiveness of Soft Wing loss in terms of the NME (\%).}
	\label{tab:loss}
\end{table}

\section{Conclusion}
In this paper, we propose a fast and accurate face alignment method.
We present a structure coherence component which consists of attention guided multi-scale feature fusion, mapping to node, dynamic adjacency matrix weighting module and graph relation network.
We utilize the relation among facial parts appropriately, which permits precise localization of facial landmarks under hard cases.
Experimental results in three challenging face alignment benchmarks demonstrate the effectiveness of the proposed method.

\clearpage
%
%
\bibliographystyle{splncs04}
\bibliography{egbib}
\end{document}